# Mothership: A Two-Tier Autonomous Refuse Collection System for Pedestrian Campuses

Wei Wang*, Yiru Veronika Wang, Yihan Wang, Zihao Li, Xiaohui Liang
Robostreet research team. Cambridge, MA 02142 USA

*Corresponding author, Email: weiwang9@mit.edu

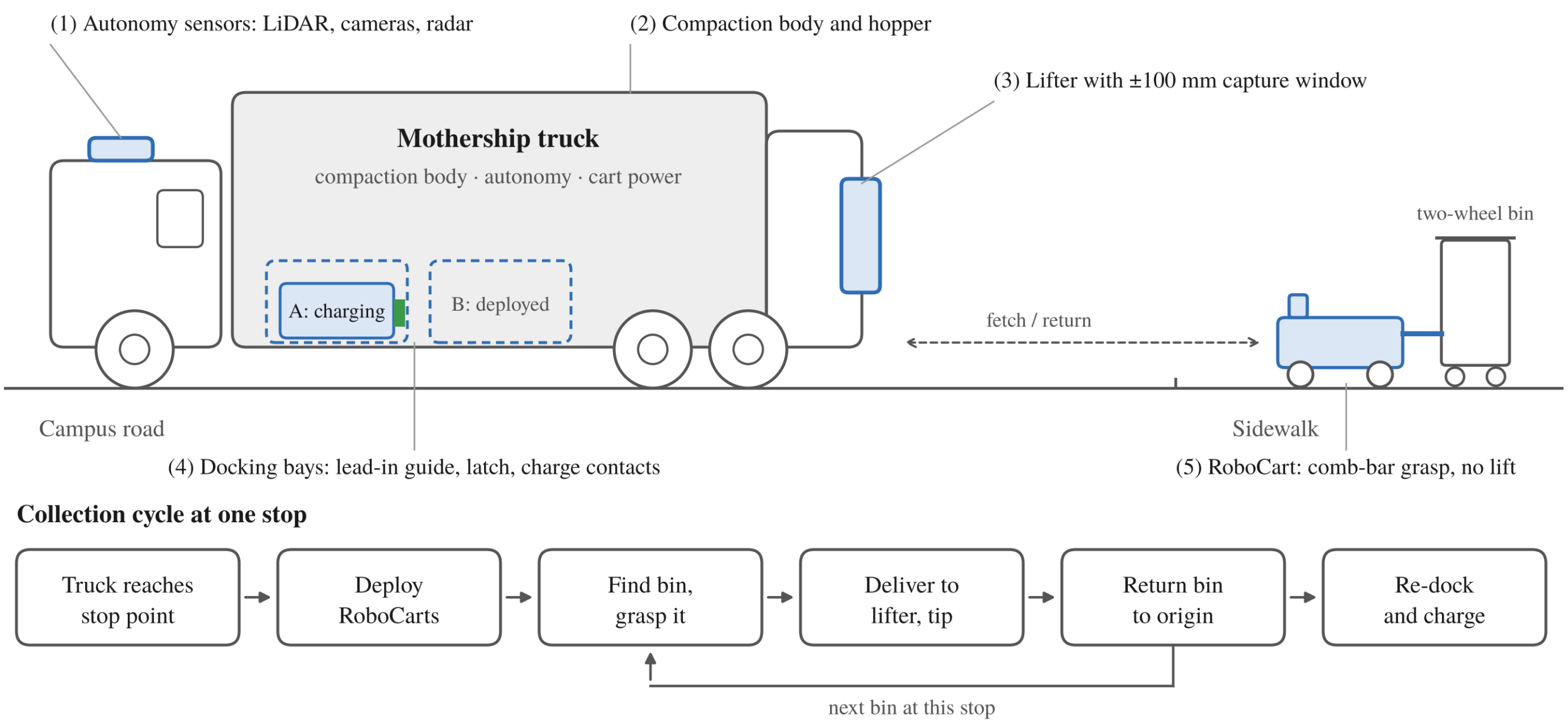

Fig. 1. Mothership concept. The truck parks at a stop point on a campus road and deploys two RoboCarts from side-mounted docking bays. Each cart fetches a bin from the sidewalk, delivers it to the truck's lifter, returns the empty bin to its original position, and re-docks to charge once the stop is cleared.

***Abstract*—Refuse collection in pedestrian campuses, parks and mixed-use districts remains manual because compaction trucks cannot reach bins located on walkways, courtyards and building service doors. This paper presents Mothership, a two-tier autonomous refuse collection system in which an electric compaction truck (the mothership) carries two low-speed sidewalk robots (RoboCarts) that fetch standard two-wheel bins, deliver them to the truck, return the empty bins to their original positions and re-dock on the truck for charging. The architecture separates responsibilities by scale: the truck performs road driving, compaction, lifting and energy supply, while the carts perform only sidewalk transport and bin grasping. This split removes the lift-and-tip mechanism from the sidewalk vehicle, keeping it light and low, and relaxes cart-to-truck docking tolerance to the ±100 mm capture window of a standard automated lifter. We describe the system architecture, the bin-handling and docking interfaces, and a mission-planning formulation that decomposes a route into truck stop points and cart tasks. A discrete-event simulation over synthetic campus layouts with 2–16 bins/ha quantifies throughput as a function of bin density, cart count and stop-point radius. A two-cart configuration serves 20–29 bins/h. Throughput is governed by an interaction between cart count and stop-point radius: a single cart performs best with many close stops, whereas three or four carts need wide stops (80 m) to stay busy, reaching 35–49 bins/h at moderate-to-high densities. The best stop radius therefore has to be chosen together with the fleet size, not fixed in advance.**

## I. Introduction

Municipal refuse collection has been progressively automated at the curb: automated side loaders (ASLs) with hydraulic arms empty a cart in 8–12 s and let a single operator serve more than a thousand households per day [1], [2]. This automation stops at the curb. In university campuses, business parks, hospital grounds and pedestrian-only districts, bins stand on walkways, in courtyards and at service doors that a Class 8 truck cannot approach, and the last 20–150 m between the bin and the truck are still covered by a worker pushing the bin by hand. Refuse and recyclable material collection remains among the five most dangerous occupations in the United States, with a 2024 fatal injury rate of 37.4 per 100,000 full-time workers [3].

Two lines of technology now exist that, taken together, could close this gap. First, low-speed autonomous driving for sanitation vehicles has reached commercial operation:

driverless sweepers operate in several cities in China, Singapore and the Middle East [4], [5]. Second, sidewalk robots have accumulated more than ten million autonomous deliveries and 22 million km on public footpaths [6]. What has not been demonstrated is a system that combines the two so that an autonomous truck and autonomous sidewalk robots jointly collect bins without human handling. The closest precedent, the Volvo Group ROAR project of 2016 [7], carried a single robot on a human-driven truck and did not progress beyond a prototype.

This paper presents Mothership, a two-tier system built around a design principle we call *split responsibility by scale*: every function that requires power, mass or reach (road driving, compaction, lifting, energy storage) is assigned to the truck; every function that requires access to narrow spaces (sidewalk transport, bin grasping, bin return) is assigned to the small robots. The principle has three consequences that distinguish Mothership from ROAR and from truck-and-robot delivery concepts [8]: (i) the sidewalk robot carries no lift mechanism, so it stays light and low; (ii) the robot-to-truck handover tolerance is absorbed by the truck's lifter capture window, so the robot needs only centimetre-level, not millimetre-level, positioning at the truck; and (iii) the robots are powered, transported and charged by the truck, so no fixed sidewalk infrastructure is required.

The contributions of this paper are:

1) A system architecture for truck-carried autonomous bin-collection robots, including the on-truck docking bay and the split-responsibility tipping interface (Sections III–IV).
2) A mission-planning formulation that decomposes a collection route into truck stop points and cart tasks, with a coverage model for stop-point selection (Section V).
3) A simulation study that quantifies system throughput against bin density, cart count and stop-point radius, and shows that the best stop-point radius grows with the number of carts (Section VI).

Section II reviews related work. Section III describes the system and its operating cycle. Section IV details bin handling and docking. Section V formulates mission planning. Section VI reports the simulation study, and Section VII concludes.

## II. Related Work

### A. Automated and Autonomous Refuse Vehicles

Curbside automation is mature. Automated side loaders grasp a standard cart with a hydraulic arm of roughly 2.7 m reach and complete a lift in about 8 s [1]; single-operator routes reach about 1,200 households per day [2]. These systems require the bin to be placed at the curb by the resident or a worker. Autonomous driving for refuse trucks was demonstrated by Volvo and Renova in 2017, where the truck reversed autonomously between bins along a pre-mapped route while the operator handled the bins on foot [9]. Low-speed driverless street sweepers are now in commercial service: WeRide reports its first commercial unmanned sanitation deployment in July 2024 with later operations in Singapore and Saudi Arabia [4], and Autowise reports more than 300 driverless sweepers in commercial use across Chinese cities [5]. These systems automate driving but not bin handling away from the road.

### B. Truck-Carried Robots

The ROAR (RObot-based Autonomous Refuse handling) project of Volvo Group, Renova, Chalmers University of Technology, Mälardalen University and Penn State (2015–2016) is the most direct precedent [7], [10]. A drone launched from the truck roof located bins; a single robot carried on the truck drove to each bin, brought it back, lifted it to empty it, and returned it. The truck was human-driven, the robot carried its own lift, and the project ended at the prototype stage. In parcel logistics, the Mercedes-Benz and Starship Robovan concept (2016) carried eight sidewalk robots in a van as a "mothership" [8]. Mothership differs from both in assigning the lift to the truck, in operating several carts concurrently against one shared lifter, and in formulating the truck route and cart tasks as one planning problem.

### C. Sidewalk Robots

Sidewalk delivery robots have established that low-speed autonomous navigation among pedestrians is practical at scale. Starship Technologies reports more than ten million deliveries, 22 million autonomous kilometres and about 125,000 road crossings per day, at a cruising speed of about 6 km/h with robots of roughly 36–38 kg [6], [11]. A bin-handling robot must carry a payload an order of magnitude heavier (a full 240 L bin weighs up to about 110 kg [12]), which raises braking distance and contact energy and motivates the speed zoning in Section IV-D. Safety requirements for robots operating near the public are addressed by ISO 13482 [13]; protective-field and speed-reduction logic for driverless carts is specified in ISO 3691-4 [14].

### D. Precision Docking

Industrial autonomous mobile robots dock to fiducial markers with a repeatability of ±2–3 mm and ±0.5°, compared with ±60–85 mm under free navigation [15]. Mothership uses fiducial docking for the on-truck bays, where electrical contacts require it, but deliberately avoids depending on it at the tipping interface, where a mechanical capture window absorbs the error (Section IV-B).

## III. System Architecture

### A. Components

The system has three tiers (Fig. 1).

**Mothership truck.** A battery-electric compaction truck with a drive-by-wire chassis and an autonomy stack (roof LiDAR, surround cameras, radar) for low-speed operation on campus roads. It carries the compaction body, a single automated lifter at the hopper, two docking bays, and a local wireless network for the carts. The truck is the only vehicle that leaves the campus.

**RoboCarts.** Two (or more) low-speed sidewalk robots, each sized to carry one full two-wheel bin. A cart has a four-wheel drive base, a comb-bar grasp mechanism with an 80 mm lift stroke, a 360° short-range LiDAR, surround RGB-D cameras, contact bumpers, and an independent safety controller. It carries no tipping mechanism.

**Fleet server.** A cloud service that maintains the bin map (position, orientation and service schedule of every bin), plans truck routes and stop points, assigns bins to carts, and provides remote supervision and teleoperation.

### B. Operating Cycle

One route proceeds as follows.

1) **Plan.** The server selects stop points that cover all scheduled bins and orders them into a truck tour (Section V).
2) **Stop and deploy.** The truck drives to the next stop, opens the bays and releases the carts.
3) **Fetch.** Each cart drives to an assigned bin, detects it, estimates its pose, records its original position and orientation, and grasps it by the comb bar.
4) **Deliver and tip.** The cart carries the bin to the handover zone at the lifter and releases it. The truck's lifter grasps, tips and lowers the bin, and the cart re-grasps it.
5) **Return.** The cart returns the empty bin to the recorded origin.
6) **Re-dock.** When the stop's bins are served, the carts return to the bays, latch and begin charging. The truck proceeds to the next stop.
7) **Unload.** When the body is full or the route is complete, the truck drives to the transfer station.

The lifter is the only resource shared by the carts; steps 3–5 for different carts run concurrently.

## IV. Bin Handling and Docking

### A. Bin Interface

Mothership targets the standard two-wheel bin defined by EN 840-1 [12] and ANSI Z245.60 Type B [16], which carries a comb bar below the lid hinge for automated lifters. A typical 240 L bin is about 1,080 mm tall, 580 mm wide and 740 mm deep, weighs 11–14 kg empty and has a rated load of 96 kg. The cart's grasp is designed for a total of 120 kg.

The cart grasps the bin from the front in four steps: it reverses to the bin face, lowers a grasp plate to comb-bar height and advances 100 mm to engage the bar; two side arms close on the bin body; the plate rises by 60–80 mm so that the bin wheels clear the ground and the bin tilts back about 10° towards the cart, placing the combined centre of mass inside the cart's wheelbase. This requires about 1.5 kN of lift over an 80 mm stroke, which a single electric linear actuator provides.

### B. Split-Responsibility Tipping

In ROAR [7] the robot lifted and tipped the bin itself. That design places a 1.5 m, 135° lift-and-tip mechanism on the sidewalk vehicle, which raises its mass, height and centre of gravity, and requires outriggers against overturning during the tip. Mothership instead leaves tipping to the truck. The cart sets the bin down in a handover zone in front of the truck's automated lifter and backs off; the lifter, whose gripper opening captures a bin within about ±100 mm of its nominal position, grasps, tips and lowers the bin; the cart re-grasps it.

The effect is on the docking error budget. If the cart had to engage a fixed hopper interface, the handover would require fiducial-level accuracy (millimetres). With the lifter's capture window, the cart's position error only needs to fall well inside ±100 mm. Free navigation on a mapped surface typically achieves ±60–85 mm [15]; adding a fiducial on the truck at the handover zone brings the cart within a few centimetres, a margin of roughly 3× over the required tolerance without contact guides.

### C. On-Truck Docking Bays

The bays occupy the space between the axles normally used for tanks and tool boxes. Each bay has:

- a deployable ramp with a slope of at most 10°;
- a V-shaped lead-in guide that accepts the cart with ±50 mm lateral error and mechanically centres it to ±5 mm;
- an electric latch, normally locked, rated for the truck's braking loads;
- spring-loaded charging contacts (inductive charging is an alternative that tolerates larger misalignment and has no exposed contacts);
- a fiducial marker and retro-reflective strip for the final 2 m of the approach.

The bay is the one place where millimetre-level alignment matters, and it is achieved mechanically by the lead-in guide rather than by the cart's navigation.

### D. Speed Zoning

A loaded cart with a total mass of about 250 kg has five to seven times the mass of a delivery robot. Contact energy, not mass alone, is what must be limited near pedestrians. Sidewalk delivery robots are reported to have a maximum impact energy of about 52 J [11]. For a 250 kg cart the speed that gives the same energy is

$$v_{\max} = \sqrt{\frac{2E}{m}} = \sqrt{\frac{2\cdot 52\ \mathrm{J}}{250\ \mathrm{kg}}} \approx 0.64\ \mathrm{m/s} \quad (1)$$

or about 2.3 km/h. Mothership therefore uses three speed zones, enforced by the cart's independent safety controller from LiDAR protective fields in the manner of ISO 3691-4 [14]: 6 km/h empty and 4 km/h loaded with no pedestrian within 3 m; 2 km/h with a pedestrian inside 3 m; stop with a pedestrian inside 1 m. The simulation in Section VI uses the unobstructed speeds and therefore gives an upper bound on throughput.

## V. Mission Planning

A route is planned in two levels: the truck level chooses and orders stop points; the cart level assigns the bins of each stop to carts.

### A. Stop-Point Selection

Let $B$ be the set of bins scheduled for service, and $P$ a set of candidate stop points sampled along the campus roads. A stop $p$ covers bin $b$ if the walking distance satisfies $d(p, b) \le R$, where $R$ is the stop-point radius. With binary variables $x_p$, stop selection is the set-cover problem

$$\min_{x} \sum_{p\in P} x_p \quad (2)$$

$$\text{s.t.}\ \sum_{p:\, d(p,b)\le R} x_p \ge 1\ \ \forall b \in B \quad (3)$$

Each bin is then assigned to one selected stop. $R$ trades off two costs: a small $R$ gives short cart trips but many stops, each carrying a fixed deployment overhead and a truck move; a large $R$ gives few stops but long cart trips.

### B. Service Time at a Stop

For bin $i$ at walking distance $d_i$ from its stop, with empty and loaded cart speeds $v_e$ and $v_l$, grasp time $t_g$, lifter handover time $t_h$ and release time $t_r$, the cart cycle is

$$c_i = \frac{2d_i}{v_e} + \frac{2d_i}{v_l} + t_g + t_h + t_r \quad (4)$$

With $k$ carts and a single lifter, the time to clear a stop with bin set $S$ of $n_S$ bins is bounded below by

$$T(S) \geq \max\left(\frac{1}{k}\sum_{i\in S} c_i\,, n_S t_h, \max_{i\in S} c_i\right) \quad (5)$$

The three terms correspond to the carts being the bottleneck, the lifter being the bottleneck, and a single distant bin setting the critical path. The last term explains why extra carts help little when a stop holds only a few bins.

### C. Route Objective

The total route time is the truck travel time plus, at each stop, a fixed deployment time $t_{\text{dep}}$ and the stop's service time:

$$T_{\text{route}} = \sum_{j=1}^{m}\left(\frac{\ell_{j-1,j}}{v_t} + t_{\text{dep}} + T(S_j)\right) \quad (6)$$

where $\ell_{j-1,j}$ is the road distance between consecutive stops and $v_t$ the truck speed on campus roads. Throughput is $|B|/T_{\text{route}}$. Joint optimization of stops, tour and cart schedules is NP-hard. In this paper we use three standard heuristics: greedy set cover for stop selection, a nearest-neighbour tour for the truck, and longest-processing-time-first list scheduling for carts at each stop, with the lifter modelled as a single-server queue.

## VI. Simulation Study

### A. Setup

We built a discrete-event simulation of the operating cycle in Section III-B using the planning heuristics of Section V. Each layout is a 600 m × 600 m campus (36 ha) with parallel roads every 150 m in both directions, so no point is more than 75 m (Manhattan) from a road. Bins are placed uniformly at random at densities of 2, 4, 8 and 16 bins/ha (72–576 bins). Candidate stop points are sampled every 10 m along roads. Walking distances are Manhattan distances. Each configuration is run on 10 random layouts. Table I lists the timing parameters, which follow Sections IV and V and the delivery-robot speeds in [6].

TABLE I
SIMULATION PARAMETERS

| Parameter | Value |
|---|---|
| Cart speed, empty / loaded | 6 / 4 km/h |
| Grasp time $t_g$ | 20 s |
| Lifter handover time $t_h$ (shared) | 25 s |
| Release time $t_r$ | 10 s |
| Deployment time $t_{\text{dep}}$ per stop | 60 s |
| Truck speed between stops $v_t$ | 15 km/h |
| Stop-point radius $R$ | 30, 50, 80 m |
| Carts per truck $k$ | 1, 2, 3, 4 |

The model omits pedestrian slow-downs, failed grasps, bin-body unloading trips and charging stops, so it gives an upper bound on throughput for the stated speeds. With $t_h$ = 25 s the lifter can serve at most 144 bins/h, and it was not the bottleneck in any configuration tested.

### B. Results

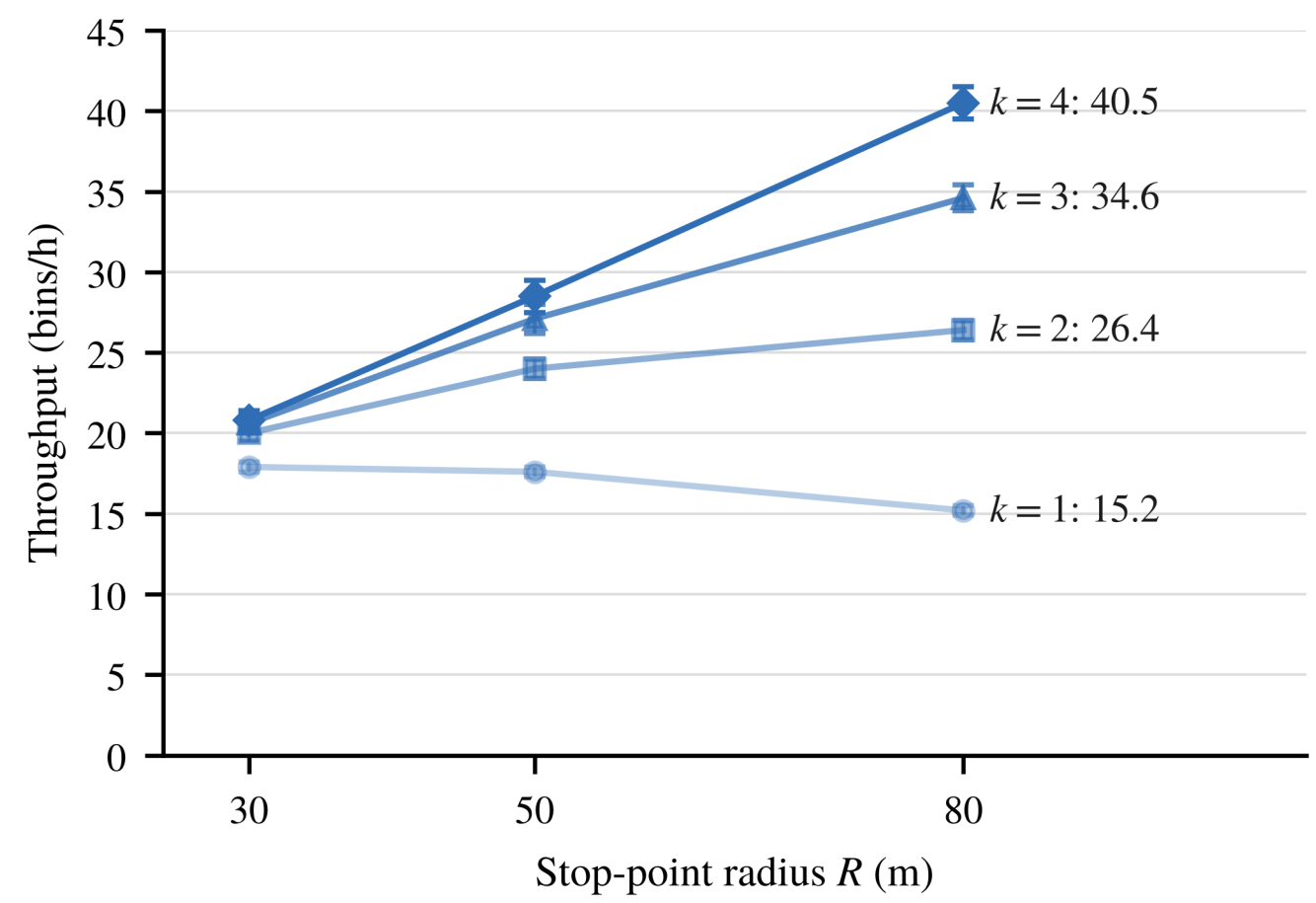

Fig. 2. Simulated throughput against stop-point radius for $k$ = 1–4 carts at 8 bins/ha. Mean of 10 random layouts; error bars show ±1 s.d.

Fig. 2 shows the central result. Throughput depends on the interaction of cart count and stop-point radius, not on either alone. With one cart, the 30 m radius performs best: a single cart cannot overlap trips, so long trips cost directly. With two or more carts the order reverses, and the 80 m radius performs best. At 30 m, extra carts add almost nothing (20.0 to 20.8 bins/h from two to four carts), because each stop holds only one or two bins and the third term of (5) dominates: carts wait for the farthest bin rather than working in parallel.

TABLE II
THROUGHPUT (BINS/H) AT THE BEST RADIUS TESTED

| Density (bins/ha) | Bins | $k$ = 1 | $k$ = 2 | $k$ = 3 | $k$ = 4 |
|---|---|---|---|---|---|
| 2 | 72 | 16.0 | 20.0 | 23.2 | 24.9 |
| 4 | 144 | 17.0 | 23.3 | 28.9 | 32.5 |
| 8 | 288 | 17.9 | 26.4 | 34.6 | 40.5 |
| 16 | 576 | 18.8 | 29.2 | 40.0 | 48.9 |

Best radius: $R$ = 30 m for $k$ = 1 and $R$ = 80 m for $k \geq 2$ at every density. Standard deviations across layouts are at most 1.4 bins/h.

Table II shows how this extends across densities. The best radius was 30 m for one cart and 80 m for two to four carts at every density. Higher density raises throughput for every fleet size because each stop holds more bins, but the gain grows with the number of carts: from 2 to 16 bins/ha, one cart improves by 18% and four carts by 96%. The two-cart configuration, which fits within the truck's side bays, delivers 20–29 bins/h. A third cart adds 16–37% over two carts, the larger gains at higher density.

### C. Discussion

Three design implications follow. First, the stop-point radius should not be fixed by policy but chosen together with the fleet size, since the best value moves from short to long as carts are added. Second, because 80 m was the largest radius tested and still best for two or more carts, the true optimum for larger fleets may lie beyond it; it is bounded in practice by the speed zoning of Section IV-D, which lengthens long trips through pedestrian areas. Third, sensitivity runs at 8 bins/ha with $t_{dep}$ varied from 20 s to 120 s shifted throughput by up to 17% but left 80 m the best radius for two or more carts. For one cart, the best radius moved from 30 m to 50 m at $t_{dep}$ = 120 s (15.7 vs. 14.8 bins/h), as expected when each extra stop costs more. The reversal between one cart and several carts persisted in every case, so it is a property of trip geometry rather than of deployment overhead.

The simulation has four limitations. The layouts are synthetic and uniform, whereas real campuses cluster bins at building entrances, which would favour larger radii further. The model assumes unobstructed speeds and no failed grasps. Cart energy is not modelled. Finally, the heuristics are not optimal, so the results are achievable lower bounds for the given model rather than optima.

## VII. Conclusion and Future Work

This paper presented Mothership, a two-tier autonomous refuse collection system in which an electric compaction truck carries, powers and coordinates low-speed sidewalk robots that fetch, deliver and return standard two-wheel bins. Its central design choice is to split responsibility by scale: lifting, compaction, energy storage and road driving stay on the truck, and the sidewalk robots only transport and grasp. This removes the lift-and-tip mechanism from the sidewalk vehicle and lets the truck's lifter capture window absorb handover error, so millimetre-level alignment is needed only inside the docking bays, where a mechanical guide provides it.

We formulated route planning as stop-point selection, truck touring and per-stop cart scheduling against a single shared lifter. Simulation over synthetic campus layouts showed that throughput is governed by the interaction between cart count and stop-point radius: one cart performs best with close stops, while two to four carts need wide stops to work in parallel. A two-cart configuration delivered 20–29 bins/h and four carts up to 49 bins/h at the radii tested.

Future work will proceed in three directions: (i) planning on real campus bin maps with clustered bins and pedestrian-flow-dependent speed zones, with joint optimization of stop points and cart schedules; (ii) a prototype RoboCart and docking bay to measure grasp success rate, handover error and docking repeatability outdoors; and (iii) a safety case for the cart under ANSI/UL 4600, with protective-field design validated against ISO 3691-4 and ISO 13482.

## References

[1] Heil Co., "DuraPack Python automated side loader," product page. [Online]. Available: https://www.heil.com/bodies/durapack-python-automated-side-load-garbage-trucks/

[2] Curotto-Can, "Advantages," product page. [Online]. Available: https://www.thecurottocan.com/advantages/

[3] Solid Waste Association of North America, statement on the 2024 BLS Census of Fatal Occupational Injuries for refuse and recyclable material collectors, Feb. 19, 2026, reported in *Waste Dive*. [Online]. Available: https://www.wastedive.com/

[4] WeRide Inc., Form F-1 registration statement, U.S. Securities and Exchange Commission, 2025. [Online]. Available: https://www.sec.gov/Archives/edgar/data/1867729/000110465925102618/wrd-20250630xf1.htm

[5] "Autowise unmanned sanitation vehicles enter large-scale commercial operation" (in Chinese), *Sina Finance*, Feb. 21, 2025. [Online]. Available: https://finance.sina.com.cn/jjxw/2025-02-21/doc-inemfsee3613164.shtml

[6] Starship Technologies, "Autonomous delivery moves into the mainstream as Starship Technologies passes 10 million deliveries," press release, Apr. 2026. [Online]. Available: https://www.starship.xyz/press/

[7] Volvo Group, "Drone to help refuse-collecting robot find refuse bins," news release, Feb. 2016. [Online]. Available: https://www.volvogroup.com/en/news-and-media/news/2016/feb/drone-to-help-refuse-collecting-robot-find-refuse-bins.html

[8] "Starship delivery robots find mothership in Mercedes Robovan," *The Robot Report*, Sep. 2016. [Online]. Available: https://www.therobotreport.com/starship-delivery-robots-find-mothership-in-mercedes-robovan/

[9] Volvo Group, "Volvo pioneers autonomous, self-driving refuse truck in the urban environment," news release, May 17, 2017. [Online]. Available: https://www.volvogroup.com/en/news-and-media/news/2017/may/news-2561936.html

[10] Penn State, "Robots to the refuse," *Research Matters*, Jun. 10, 2016. [Online]. Available: https://researchmatters.psu.edu/2016/06/10/robots-to-the-refuse

[11] Starship Technologies, "Frequently asked questions." [Online]. Available: https://www.starship.xyz/faq/

[12] *Mobile Waste and Recycling Containers—Part 1: Containers with 2 Wheels with a Capacity up to 400 l for Comb Lifting Devices—Dimensions and Design*, EN 840-1:2020, CEN, 2020.

[13] *Robots and Robotic Devices—Safety Requirements for Personal Care Robots*, ISO 13482:2014, ISO, 2014.

[14] *Industrial Trucks—Safety Requirements and Verification—Part 4: Driverless Industrial Trucks and Their Systems*, ISO 3691-4:2023, ISO, 2023.

[15] Mobile Industrial Robots, "MiR250 specifications," rev. 2.81. [Online]. Available: https://shelley.com/wp-content/uploads/2024/07/MiR250-Specifications-2.81.pdf

[16] *Mobile Wastes and Recyclable Materials Collection, Transportation and Compaction Equipment and Facilities—Waste Containers—Compatibility Dimensions*, ANSI Z245.60-2008, ANSI, 2008.